\documentclass[a4paper,fleqn]{cas-sc}

\usepackage[authoryear]{natbib}
\usepackage{pifont}
\usepackage{enumitem}
\usepackage{array}  

\def\tsc#1{\csdef{#1}{\textsc{\lowercase{#1}}\xspace}}
\tsc{WGM}
\tsc{QE}
\tsc{EP}
\tsc{PMS}
\tsc{BEC}
\tsc{DE}

\begin{document}
\let\WriteBookmarks\relax
\def\floatpagepagefraction{1}
\def\textpagefraction{.001}

\shorttitle{VLM-Based Safe End-to-End Cooperative Autonomous Driving with Long-Tail Modeling}

\shortauthors{Junwei You et~al.}

\title [mode = title]{\textbf{SEAL}: Vision-Language Model-Based \textbf{S}afe \textbf{E}nd-to-End Cooperative \textbf{A}utonomous Driving with Adaptive \textbf{L}ong-Tail Modeling}                      

\author[1]{Junwei You}[orcid=0009-0002-6447-8276]
                        % [type=editor,
                        % auid=000,bioid=1]
                        % prefix=Sir]
                        % role=Researcher,
                        % orcid=0000-0001-7511-2910]

% \cormark[1]


\ead{jyou38@wisc.edu}

% \ead[url]{www.cvr.cc, cvr@sayahna.org}

% \credit{Conceptualization of this study, Methodology, Software}

\affiliation[1]{organization={Department of Civil and Environmental Engineering, University of Wisconsin–Madison},
    % addressline={Radarweg 29}, 
    city={Madison},
    state={WI},
    % citysep={}, % Uncomment if no comma needed between city and postcode
    postcode={53706}, 
    country={USA}}

\affiliation[2]{organization={College of Computing and Data Science, Nanyang Technological University},
    % addressline={Radarweg 29}, 
    city={Singapore},
    % state={WI},
    % citysep={}, % Uncomment if no comma needed between city and postcode
    postcode={639798}, 
    country={Singapore}}

\affiliation[3]{organization={College of Transportation, Tongji University},
    city={Shanghai},
    postcode={201804}, 
    country={China}}

\author[1]{Pei Li}
\ead{pei.li@wisc.edu}
\cormark[1]

\author[2]{Zhuoyu Jiang}
\ead{JI0007YU@e.ntu.edu.sg}

\author[1]{Zilin Huang}
\ead{zilin.huang@wisc.edu}

\author[1]{Rui Gan}
\ead{rgan6@wisc.edu}

\author[3,1]{Haotian Shi}
\ead{shihaotian95@tongji.edu.cn}

\author[1]{Bin Ran}
\ead{bran@wisc.edu}

% \credit{Data curation, Writing - Original draft preparation}

\cortext[cor1]{Corresponding author}
% \cortext[cor2]{Principal corresponding author}

% \fntext[fn1]{This is the first author footnote. but is common to third
% \fntext[fn2]{Another author footnote, this is a very long footnote and
%   it should be a really long footnote. But this footnote is not yet
%   sufficiently long enough to make two lines of footnote text.}

% \nonumnote{This note has no numbers. In this work we demonstrate $a_b$
%   the formation Y\_1 of a new type of polariton on the interface
%   between a cuprous oxide slab and a polystyrene micro-sphere placed


\begin{abstract}
Autonomous driving technologies face significant safety challenges while operating under rare, diverse, and visually degraded weather scenarios. These challenges become more critical in cooperative settings, where vehicles and infrastructure jointly perceive and reason across complex environments. To address these issues, we propose SEAL, a vision-language model-based framework with adaptive multimodal learning for robust cooperative autonomous driving under long-tail scenarios. SEAL introduces three core innovations: (i) a \textit{prompt-driven long-tail scenario generation and evaluation pipeline} that leverages foundation models to synthesize realistic long-tail conditions such as snow and fog across vehicle- and infrastructure-side views, enriching training diversity efficiently; (ii) a \textit{gated multi-scenario adaptive attention} module that modulates the visual stream using scenario priors to recalibrate ambiguous or corrupted features; and (iii) a \textit{multi-task scenario-aware contrastive learning} objective that improves multimodal alignment and promotes cross-scenario feature separability. Extensive experiments demonstrate that SEAL significantly outperforms existing baselines in reasoning, safety, and planning accuracy under complex, challenging driving conditions, advancing the safety, robustness, and scalability of autonomous driving.
\end{abstract}

% Keywords
\begin{keywords}
End-to-end autonomous driving \sep V2X cooperation \sep Vision-language model \sep Autonomous driving safety \sep Trajectory planning

\end{keywords}
\maketitle

\section{Introduction}

Autonomous driving has the potential to revolutionize transportation systems by operating with minimal or no human intervention, thereby eliminating many human errors. However, it faces significant safety challenges in diverse traffic scenarios, extreme weather conditions, and complex interactions with human-driven vehicles (\cite{zhou2023identifying,XU2025108139}). Existing studies have suggested that extreme weather conditions, including heavy rain, snow, and fog, can significantly reduce the ability of autonomous driving in detecting objects, planning trajectories, and making driving decisions (\cite{zang2019impact}, \cite{mehra2020reviewnet}), posing significant safety concerns. Therefore, there is an urgent need to develop autonomous driving technologies that are robust to rare and diverse environmental conditions to ensure safe and reliable operation.

\begin{figure}[t]
  \centering
  \includegraphics[width=0.8\linewidth]{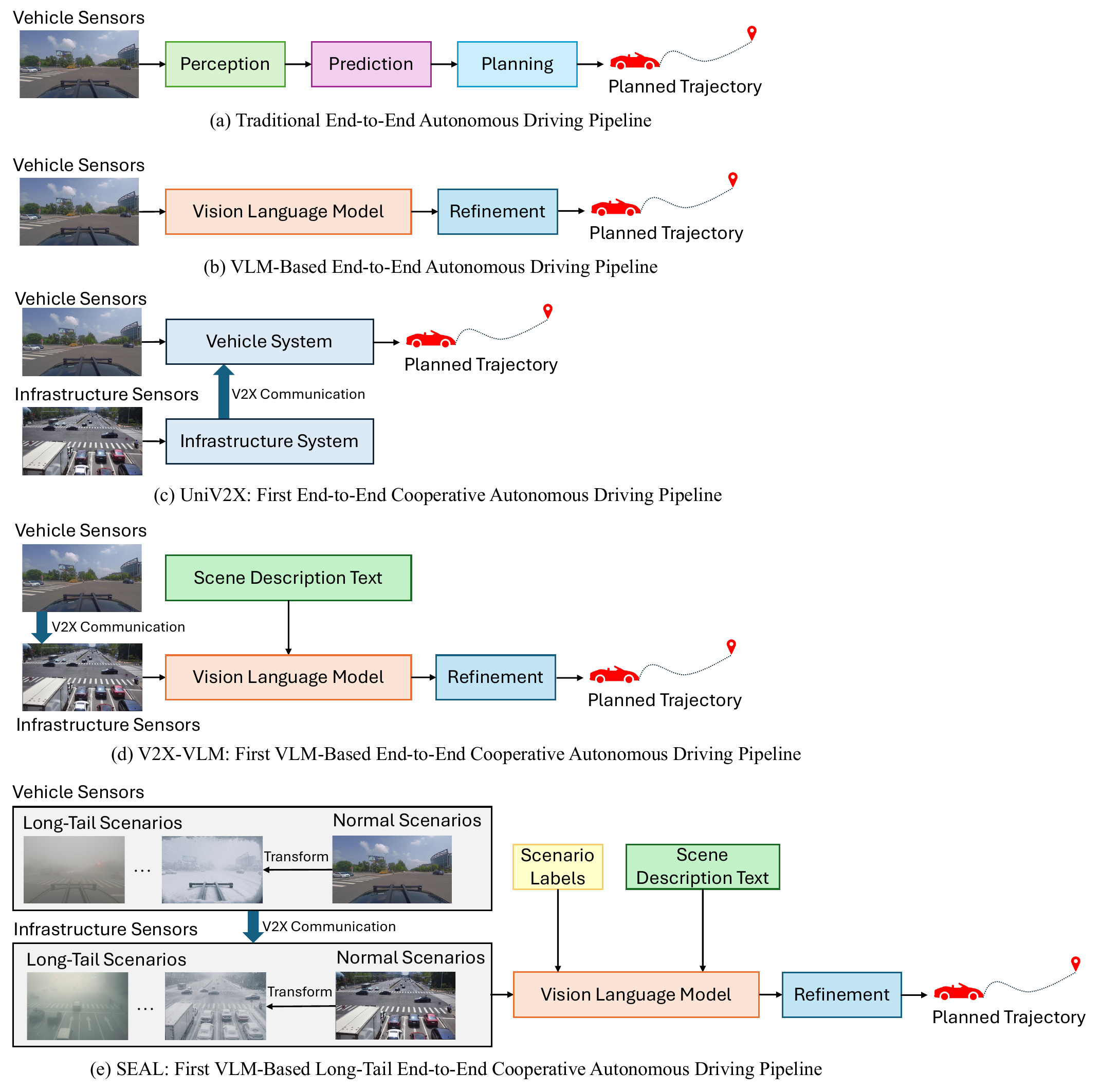}
  \caption{Overview of representative end-to-end autonomous driving pipelines. (a) Traditional pipeline separates perception, prediction, and planning. (b) VLM-based framework integrates semantic reasoning but lacks cross-agent collaboration. (c) UniV2X (\cite{yu2025end}) introduces cooperative planning without semantic alignment. (d) V2X-VLM (\cite{you2024v2x}) introduces multimodal fusion but lacks long-tail robustness. (e) SEAL integrates VLM, scenario awareness, and long-tail adaptation.}
  \label{fig:pipeline}
\end{figure}

End-to-end autonomous driving has become the prevailing paradigm for high-level autonomy, offering a unified learning-based framework that directly maps raw sensory inputs to driving actions. Compared to traditional modular pipelines that decompose driving into discrete stages~(Figure~\ref{fig:pipeline}a), end-to-end frameworks simplify system design, reduce interface mismatches, and enable global optimization of the driving policy. Among recent advances, vision-language models (VLMs) have emerged as powerful backbones for end-to-end reasoning (\cite{xiao2024florence, chen2024spatialvlm, feng2025verdi}). By jointly encoding visual scenes and natural language prompts, VLMs offer rich semantic grounding and flexible context understanding, making them a natural fit for autonomous driving tasks. Leveraging these capabilities, recent works have introduced VLM-based end-to-end pipelines (Figure~\ref{fig:pipeline}b), which show promising improvements in semantic generalization and reasoning ability (\cite{hwang2024emma, xing2025openemma, zhou2025opendrivevla, huang2024vlm, tian2024drivevlm}).

However, existing VLM-based frameworks typically operate in a single-agent setting, relying solely on vehicle-side views. This limits their spatial awareness and makes them vulnerable in occluded or visually degraded environments. To overcome these limitations, cooperative autonomous driving has gained traction by enabling collaboration between vehicles and infrastructure. Recent frameworks such as UniV2X (\cite{yu2025end}) (Figure~\ref{fig:pipeline}c) proposed the first end-to-end cooperative pipeline, directly generating trajectories from synchronized multi-agent sensory inputs. By incorporating vehicle-to-everything (V2X) communication, UniV2X improves spatial awareness and scene coverage. Moreover, \cite{you2024v2x}  proposed the first VLM-based cooperative end-to-end autonomous driving framework (Figure~\ref{fig:pipeline}d). By integrating vehicle- and infrastructure-side visual inputs with natural language scene descriptions, V2X-VLM exploits the semantic grounding, cross-modal alignment, and reasoning capabilities of large VLMs. This enables more precise, interpretable, and context-aware trajectory planning, while significantly enhancing multimodal fusion compared to earlier cooperative designs.

% Nonetheless, its architecture lacks semantic abstraction and primarily relies on black-box feature aggregation, which hampers generalization across diverse scenarios and limits interpretability. Moreover, UniV2X lacks the capacity for high-level contextual reasoning, as it does not incorporate explicit semantic priors or structured scene understanding.

% To address these challenges, our prior work V2X-VLM (\cite{you2024v2x}) While V2X-VLM has demonstrated promising results in normal scenarios and relatively 

Most existing studies have focused on developing autonomous driving frameworks under normal conditions. However, their performance under more challenging and diverse domain shifts remains underexplored. This gap is critical, as the ability to generalize under rare and adverse conditions is a key safety bottleneck for the large-scale real-world deployment of high-level intelligent driving systems. Moreover, existing works, such as \cite{you2024v2x}, employ a generic alignment objective and a static perception pathway, which may limit their robustness and adaptability in the presence of significant scenario variations.

In this work, we introduce SEAL, a VLM-based robust end-to-end cooperative autonomous driving framework with adaptive long-tail modeling (Figure~\ref{fig:pipeline}e). SEAL extends V2X-VLM with three key innovations designed to enhance resilience under challenging long-tail conditions: (i) a \textit{prompt-driven long-tail scenario generation and evaluation} pipeline that leverages state-of-the-art pretrained foundation models to synthesize realistic long-tail scenarios, such as snow and fog degradations, on both vehicle-side and infrastructure-side views, effectively enriching training diversity without incurring the high costs and complexities associated with real-world long-tail data collection; (ii) a \textit{gated multi-scenario adaptive attention (GMSAA)} mechanism that integrates scenario priors into the visual processing pipeline of the VLM, dynamically modulating feature representations to mitigate noise, ambiguity, and distribution shifts under diverse and adverse driving scenarios; and (iii) a \textit{multi-task scenario-aware contrastive learning (MSCL)} objective that jointly aligns visual and language embeddings while enforcing discriminative separation across scenarios. These components equip SEAL with the robustness and semantic grounding necessary for reliable planning in adverse and long-tail environments. Our contributions are summarized as follows:
\begin{itemize}
    \item We propose SEAL, a VLM-based framework to address safety challenges in end-to-end autonomous driving under long-tail, complex weather scenarios. To overcome the scarcity and high acquisition cost of long-tail driving data, we introduce a prompt-driven pipeline for scalable and photorealistic data generation. This enables fine-grained augmentation on both the vehicle and infrastructure views.
    \item We design a GMSAA module that adaptively incorporates scenario priors via gated attention and inter-domain similarity modeling. This enhances framework robustness by discriminating and recalibrating diverse scenario features.
    \item We formulate a multi-task contrastive learning objective that unifies modality alignment with scenario-aware domain separation to achieve stronger generalization, clearer feature disentanglement, and improved interpretability.
    \item We conduct extensive experiments on real-world V2X data. Results demonstrate that SEAL significantly outperforms prior baselines across accuracy, safety, and robustness under long-tail weather scenarios.

\end{itemize}

The remainder of this paper is organized as follows. Section~\ref{sec:related} reviews related work on cooperative autonomous driving and long-tail adaptation. Section~\ref{sec:methodology} presents the SEAL framework in detail, elaborating on its overall architecture, long-tail scenario generation pipeline, adaptive attention module, and contrastive learning objectives. Section~\ref{sec:experiments} introduces the experimental setup, datasets, evaluation metrics, and reports comprehensive results including comparisons with prior baselines, ablation studies, and qualitative analyses. Finally, Section~\ref{sec:conclusion} concludes the paper and discusses future research directions.

\section{Related Work}
\label{sec:related}

\subsection{Cooperative Autonomous Driving}

Cooperative autonomous driving aims to enhance situational awareness and decision-making by enabling vehicles and infrastructure to exchange information through vehicle-to-everything (V2X) communication, including vehicle-to-vehicle (V2V) (\cite{wang2020v2vnet, feng2023wireless, xu2023v2v4real}) and vehicle-to-infrastructure (V2I) (\cite{yang2020generating, hawlader2023vehicle, mo2024enhanced}) communication. By sharing complementary observations across agents with diverse viewpoints, these systems improve detection under occlusion, extend perception coverage, and facilitate coordination in dense and dynamic environments.

To realize these benefits, classical cooperative autonomous driving approaches typically focus on specific functional modules of the autonomy stack—namely perception, prediction, and planning—under cooperative settings. A wide range of research has explored cooperative perception, where spatially aligned multi-agent feature fusion methods (\cite{chen2019cooper, xu2022v2x, cui2022coopernaut, zhao2024coopre, ma2024macp}) are developed to enhance 3D object detection in challenging settings with partial observability. Building upon shared and spatially aligned features, cooperative prediction methods (\cite{zhang2024co, zhang2025co, wang2025cmp}) aim to jointly forecast the future trajectories by modeling their spatiotemporal interactions and mutual influences, thereby enhancing prediction accuracy under dense and interactive driving scenarios. For cooperative planning and decision-making, prior works (\cite{during2016cooperative, deng2019cooperative, chen2023interactive}) have developed various frameworks to enable effective coordination and joint policy optimization among multiple intelligent agents.

While these approaches have advanced individual autonomous driving functions, they rely on modular pipelines and fixed communication protocols, which limit scalability and adaptability. To address this, a new line of work emerges to explore end-to-end cooperative autonomous driving. UniV2X (\cite{yu2025end}) represents the first framework to unify vehicle-side and infrastructure-side observations into a single end-to-end cooperative autonomous driving system. It effectively encodes multi-view sensor inputs from both agents and directly predicts future trajectories. While UniV2X demonstrates strong performance in structured environments, its model design lacks semantic modularity and relies on rigid fusion strategies, whichb limits its adaptability across different driving contexts.

To address these constraints, our prior work V2X-VLM (\cite{you2024v2x}) proposed the first VLM-based end-to-end cooperative autonomous driving architecture. By integrating scene-level natural language descriptions with vehicle-side and infrastructure-side images, V2X-VLM fine tunes the pretrained VLM to improve semantic reasoning and multimodal alignment. This allows for enhanced interpretability and broader contextual understanding compared to earlier architectures. However, both UniV2X and V2X-VLM have not been thoroughly evaluated under complex, long-tail conditions such as adverse weather conditions, which are critical for ensuring reliability and safety in large-scale real-world autonomous driving deployments.

\subsection{Long-Tail Modeling in Autonomous Driving}
\label{sec:related-lt}

Long-tail scenarios, such as adverse weather, rare object interactions, and complex edge cases, pose significant challenges to the safety and generalization of autonomous driving systems. These scenarios often fall outside the distribution of standard training data, leading to poor model reliability and increased failure rates during deployment. As high-level autonomy moves toward real-world scalability, addressing the long-tail problem becomes indispensable.

Recent efforts have explored diverse strategies for long-tail modeling. Data-centric approaches such as LTDA-Drive (\cite{yurt2025ltda}) and LoT-nuScenes (\cite{mi2024lot}) leverage generative models or virtual simulators to augment rare scenarios, while LiloDriver (\cite{yao2025lilodriver}) introduces lifelong learning mechanisms to adapt motion planning across evolving environments. Parallel vision frameworks (\cite{wang2022parallel}) simulate multiple environment variants in parallel to regularize decision boundaries and improve robustness. On the algorithmic side, risk-aware models such as RiskNet (\cite{liu2025risknet}) and graph-based reasoning frameworks (\cite{li2023graph}) explicitly encode interaction dynamics and semantic uncertainty to forecast hazards under long-tail conditions.

In planning, dynamically conservative strategies (\cite{zhou2022dynamically}) adjust the planner's behavior to account for rare, ambiguous situations. Object-level tokenization (\cite{tian2024tokenize}) further proposes to represent scenes as composable semantic units, enhancing generalization and interpretability. These works highlight the importance of both robust data generation and adaptive algorithm design in managing distribution shifts.

Despite these advances, most existing solutions focus on specific modules of the autonomy stack and lack a unified, end-to-end framework that jointly models multimodal perception, reasoning, and planning under long-tail scenarios. Our proposed SEAL addresses this gap by integrating long-tail generation, adaptive attention, and scenario-aware contrastive learning into a unified end-to-end cooperative autonomous driving paradigm.

\section{Methodology}
\label{sec:methodology}
\subsection{Overall Framework}
\label{sec:realm-overview}

SEAL is a VLM-based end-to-end cooperative autonomous driving framework designed to enhance trajectory planning robustness under long-tail scenarios such as heavy snow and dense fog. It extends the architecture of V2X-VLM (\cite{you2024v2x}) by introducing three fundamental components: (i) an efficient prompt-driven high-fidelity long-tail image transformation pipeline, (ii) a gated multi-scenario adaptive attention module (GMSAA) for scenario-aware feature calibration and explicit separation of weather-specific cues, and (iii) a multi-task scenario-aware contrastive learning (MSCL) objective that jointly aligns vision–language embeddings while enforcing inter-scenario discrimination.

\begin{figure}[t]
  \centering
  \includegraphics[width=\linewidth]{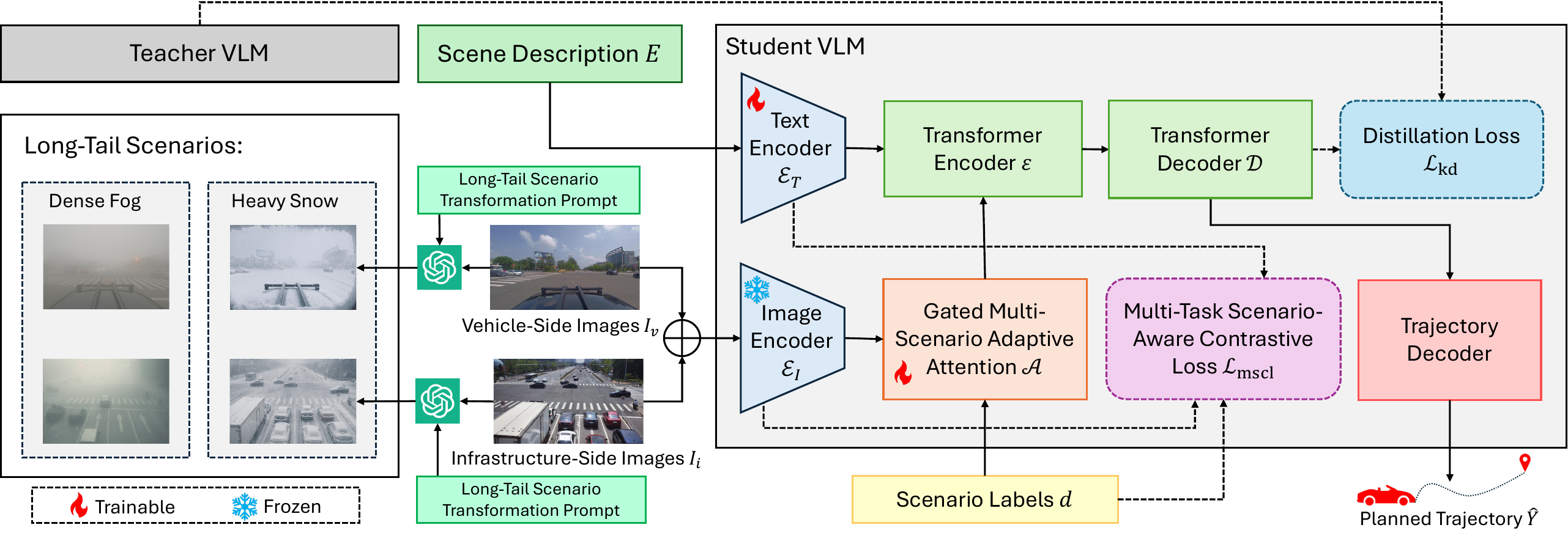}
  \caption{Overview of the SEAL framework. The vehicle-side and infrastructure-side views \(I_v, I_i\) are concatenated and processed by a frozen image encoder \(\mathcal{E}_I\) to produce visual token embeddings \(z\). The GMSAA module \(\mathcal{A}\) injects scenario-aware priors into \(z\) based on scenario label \(d\). A trainable text encoder \(\mathcal{E}_T\) encodes the scene description \(E\) into textual embeddings \(h\). The two modalities are fused via a transformer encoder \(\mathcal{E}\), and decoded by \(\mathcal{D}\) to generate a future trajectory sequence \(\hat{Y}\).}
  \label{fig:framework}
\end{figure}

Figure~\ref{fig:framework} presents the framework of SEAL. Let \(I_v, I_i \in \mathbb{R}^{3 \times H \times W}\) denote the paired vehicle- and infrastructure-side images, while \(E\) represents a natural language description of the driving scene. The two views in both the normal and long-tail scenarios generated using the prompt-based transformation method are concatenated to form a composite dual-view image, which is passed into a frozen image encoder \(\mathcal{E}_I\) to produce unified visual token embeddings \(z = \mathcal{E}_I([I_v, I_i]) \in \mathbb{R}^{N \times d}\). To introduce scenario-awareness, the extracted visual features $z$ are modulated by the GMSAA module \(\mathcal{A}\), which conditions the representation on the scenario label \(d \in \{0{:}\text{Normal}, 1{:}\text{Snow}, 2{:}\text{Fog}\}\). This mechanism injects scenario-specific priors into the visual stream, enhancing robustness under rare and adverse conditions.

Simultaneously, the text description \(E\), generated from a pre-trained large VLM and refined through human curation to accurately describe the scene, is processed by a trainable text encoder \(\mathcal{E}_T\), resulting in semantic embeddings \(h = \mathcal{E}_T(E) \in \mathbb{R}^{M \times d}\) that capture high-level contextual information. The modulated visual tokens \(\widetilde{z} = \mathcal{A}(z, d)\) and text embeddings \(h = \mathcal{E}_T(E)\) are first fused through a transformer encoder $\mathcal{E}$, which jointly processes the two modalities into a unified latent representation. This fused representation is then passed into an autoregressive transformer decoder \(\mathcal{D}\), which generates a sequence of \(K\) trajectory tokens over a vocabulary \(\mathcal{V}\):
\begin{equation}
\hat{Y} = \mathcal{D}(\mathcal{E}(\widetilde{z}, h)) \in \mathbb{R}^{K \times |\mathcal{V}|}.
\end{equation}

The training of SEAL is guided by three complementary objectives: a primary autoregressive language modeling loss \(\mathcal{L}_{\text{gen}}\) that trains the decoder to predict accurate future trajectories, a multi-task contrastive alignment loss \(\mathcal{L}_{\text{mscl}}\) that jointly enforces cross-modal consistency and inter-scenario separation, and a knowledge distillation loss \(\mathcal{L}_{\text{kd}}\) that encourages the student VLM to match the output distribution of a frozen teacher model. This integrated architecture enables SEAL to effectively leverage multimodal information and synthetic long-tail data, achieving robust trajectory planning performance under diverse and challenging conditions.

%--------------------------------------------------------
\subsection{Prompt-Driven Long-Tail Scenario Generation and Evaluation}
\label{sec:realm-longtail}

Existing cooperative driving datasets such as DAIR-V2X (\cite{yu2022dair}) and V2X-Real (\cite{xiang2024v2x}) primarily contain scenes captured under normal driving conditions, offering limited exposure to rare but safety-critical long-tail scenarios. To address this issue, this study focuses on two abnormal conditions, including heavy snow and dense fog. These weather conditions significantly degrade sensor visibility and model reliability, challenging the performance of autonomous driving. Prior works often develop dedicated generative models to generate long-tail scenarios, which require extensive tuning and domain adaptation. Instead, we adopt a prompt-driven transformation approach that leverages the recent GPT-4o (\cite{openai_gptimage1_2025}) model’s powerful image generation capabilities. This method enables efficient, high-fidelity synthesis of photorealistic snow and fog variations on existing images, with minimal overhead and reduced fine-tuning. The resulting scene transformations are not only semantically and geometrically consistent but also easily customizable through natural language prompts, making this approach scalable and scenario-controllable for long-tail data augmentation.

Given a synchronized pair of vehicle-side and infrastructure-side views \((I_v^{(n)}, I_i^{(n)})\) captured under normal conditions, we generate transformed counterparts \((\widetilde{I}_v, \widetilde{I}_i)\) using GPT-4o conditioned on weather-specific transformation prompts. The prompts are crafted to enforce detailed control over the physical attributes of snow or fog. Table~\ref{tab:prompt-snow} and Table~\ref{tab:prompt-fog} list the complete prompt design for snow and fog transformations, respectively.

\begin{table}[t]
\centering
\caption{Prompt for Snow Scene Transformation}
\label{tab:prompt-snow}
\begin{tabular}{>{\centering\arraybackslash}m{0.12\linewidth} | m{0.82\linewidth}}
\toprule
\textbf{Role} & \textbf{Content} \\
\midrule
System &
You are a professional image transformation specialist for autonomous driving scenarios. Create photorealistic snow transformations with accurate snow physics and lighting effects. \\
\midrule
User &
Please transform this image captured by the \texttt{\{ego vehicle front camera\}} and \texttt{\{infrastructure camera\}} into a realistic heavy snow scene for autonomous driving perception evaluation.

\vspace{\baselineskip} \textbf{Apply the following transformations}: 
\begin{itemize}
\item Add falling snow particles of varying sizes throughout the image; 
\item Create snow accumulation on appropriate horizontal surfaces (road edges, vehicle tops); 
\item Apply snow coverage to roadside areas, signs, and infrastructure;
\item Reduce road marking visibility with partial snow coverage on the roadway; 
\item Add appropriate brightness/reflection changes due to snow's high albedo; 
\item Create realistic snow flurry effects that partially obscure distant objects; 
\item Add snow buildup on edges of what would be the camera lens/housing;
\item Introduce glare effects where light sources interact with falling snow. 
\end{itemize}

\vspace{\baselineskip} \textbf{Technical specifications}:
\begin{itemize}
\item Maintain original image resolution and aspect ratio;
\item Ensure realistic snow physics (size, distribution, accumulation patterns); 
\item Apply proper light reflectance properties of snow surfaces; 
\item Create realistic road conditions with tire tracks where appropriate; 
\item Adjust overall scene brightness and contrast to account for snow's reflective properties;
\item Ensure snow distribution follows physical laws (more on horizontal surfaces, less on vertical); 
\item Add subtle blue tint to shadows in snow areas; 
\item Create a realistic depth effect with denser snow appearance in the distance. 
\end{itemize}

\vspace{\baselineskip} This transformed image will help expand the dataset with realistic snowy driving scenarios, dedicated to enhancing model performance in challenging winter conditions.\\
\bottomrule
\end{tabular}
\end{table}

\begin{table}[t]
\centering
\caption{Prompt for Fog Scene Transformation}
\label{tab:prompt-fog}
\begin{tabular}{>{\centering\arraybackslash}m{0.12\linewidth} | m{0.82\linewidth}}
\toprule
\textbf{Role} & \textbf{Content} \\
\midrule
System &
You are a professional image transformation specialist for autonomous driving scenarios. Create photorealistic fog transformations with accurate fog physics and lighting effects.\\
\midrule
User &
Please transform this image captured by the \texttt{\{ego vehicle front camera\}} and \texttt{\{infrastructure camera\}} into a realistic dense fog scene for autonomous driving perception evaluation.

\vspace{\baselineskip} \textbf{Apply the following transformations}: 
\begin{itemize}
\item Add realistic fog effect with visibility reduced to approximately 30-50 meters; 
\item Create a gradual fog density that increases with distance from camera; 
\item Reduce contrast and color saturation throughout the image; 
\item Add light diffusion effects around bright objects (lights, signals); 
\item Maintain the structural integrity of all key elements (vehicles, pedestrians, roads, signs); 
\item Ensure that closer objects remain more visible than distant ones; 
\item Add subtle light halos where applicable (headlights, traffic signals); 
\item Apply a slight uniform brightening effect to simulate light scattering in fog. 
\end{itemize}

\vspace{\baselineskip} \textbf{Technical specifications}:
\begin{itemize}
\item Maintain original image resolution and aspect ratio; 
\item Ensure the fog effect follows accurate atmospheric physics principles; 
\item Keep road markings partially visible but degraded according to distance; 
\item Apply appropriate fog-induced changes to shadows and reflections; 
\item Create realistic depth-dependent visibility falloff; 
\item Simulate the slight color shift typical in foggy conditions (slightly cooler tones); 
\item Add subtle volumetric lighting effects where light sources interact with fog; 
\item Ensure consistent fog density across the entire frame with proper perspective. 
\end{itemize}

\vspace{\baselineskip} This transformed image will help expand the dataset with realistic snowy driving scenarios, dedicated to enhancing model performance in challenging low-visibility conditions.\\
\bottomrule
\end{tabular}
\end{table}

To ensure spatial and semantic consistency, both vehicle and infrastructure views are transformed using a shared prompt, where we only need to specify camera perspectives. Transformed images are annotated with scenario labels \(d \in \{0{:}\text{Normal}, 1{:}\text{Snow}, 2{:}\text{Fog}\}\). Moreover, to comprehensively assess the generation quality, we employ five metrics that jointly evaluate perceptual fidelity, realism, and semantic preservation. Specifically, we use: Learned Perceptual Image Patch Similarity (LPIPS) (\cite{snell2017learning}) to measure perceptual similarity between the original and transformed images; Blind/Referenceless Image Spatial Quality Evaluator (BRISQUE) (\cite{mittal2011blind}) to quantify naturalness without needing a reference image; Fréchet Inception Distance (FID) (\cite{obukhov2020quality}) to assess distributional similarity to real samples; Fog Aware Density Evaluator (FADE) (\cite{choi2015referenceless}) to measure the degree of visibility degradation; and Semantic Intersection-over-Union (Semantic IoU) (\cite{rahman2016optimizing}) to evaluate how well the semantic structure is preserved post-transformation.

To obtain a unified and context-aware assessment of generation quality, we compute a weather-specific composite score by performing weighted fusion of normalized metric values. Given the inherent trade-offs in visual realism and structural fidelity under adverse weather conditions, not all metric scores contribute equally or monotonically to the final evaluation. To ensure interpretability, each metric is first normalized to the $[0,1]$ range. For metrics where lower raw values indicate better quality — such as LPIPS, BRISQUE, FID, and FADE — we apply inverse normalization, ensuring that higher values consistently reflect better performance. Let $m_i$ denote the raw value of the $i$-th metric, and the normalized score $s_i^{\text{w}}$ under weather condition $\text{w}$ is computed as:
\begin{equation}
s_i^{\text{w}} =
\begin{cases}
1 - \dfrac{m_i}{100}, & \text{if } i \in \{\text{BRISQUE}, \text{FID} \} \\
1 - m_i, & \text{if } i \in \{\text{LPIPS}, \text{FADE} \} \\
m_i, & \text{if } i = \text{Semantic IoU}
\end{cases}
\end{equation}

Based on these normalized scores, the composite quality score $\mathcal{S}_{\text{comp}}^{\text{w}}$ is defined as the square root of the weighted sum:
\begin{equation}
\mathcal{S}_{\text{comp}}^{\text{w}} = \left( \sum_{i=1}^{5} w_i^{\text{w}} \cdot s_i^{\text{w}} \right)^{1/2}, \quad \text{where } \sum_{i=1}^{5} w_i^{\text{w}} = 1
\end{equation}

The weights $w_i^{\text{w}}$ are adaptively assigned for each weather condition to emphasize metrics most relevant to realism. For snow, greater weight is placed on perceptual similarity and semantic structure, while for fog, visibility-aware metrics such as FADE and FID receive higher importance. This adjustment is necessary, since adverse conditions naturally lead to visual deviations like reduced contrast, which should not be penalized by standard metrics. The evaluation thus remains perceptually aligned and physically meaningful.

This weather-aware scoring pipeline provides a rigorous standard for selecting and validating augmented samples. It ensures that generated scenes used in training reflect high-quality transformations with plausible structure and degraded visibility, thereby improving downstream robustness under long-tail conditions.

\subsection{Gated Multi-Scenario Adaptive Attention}
\label{sec:gmsaa}

\begin{figure}[t]
    \centering
    \includegraphics[width=0.9\linewidth]{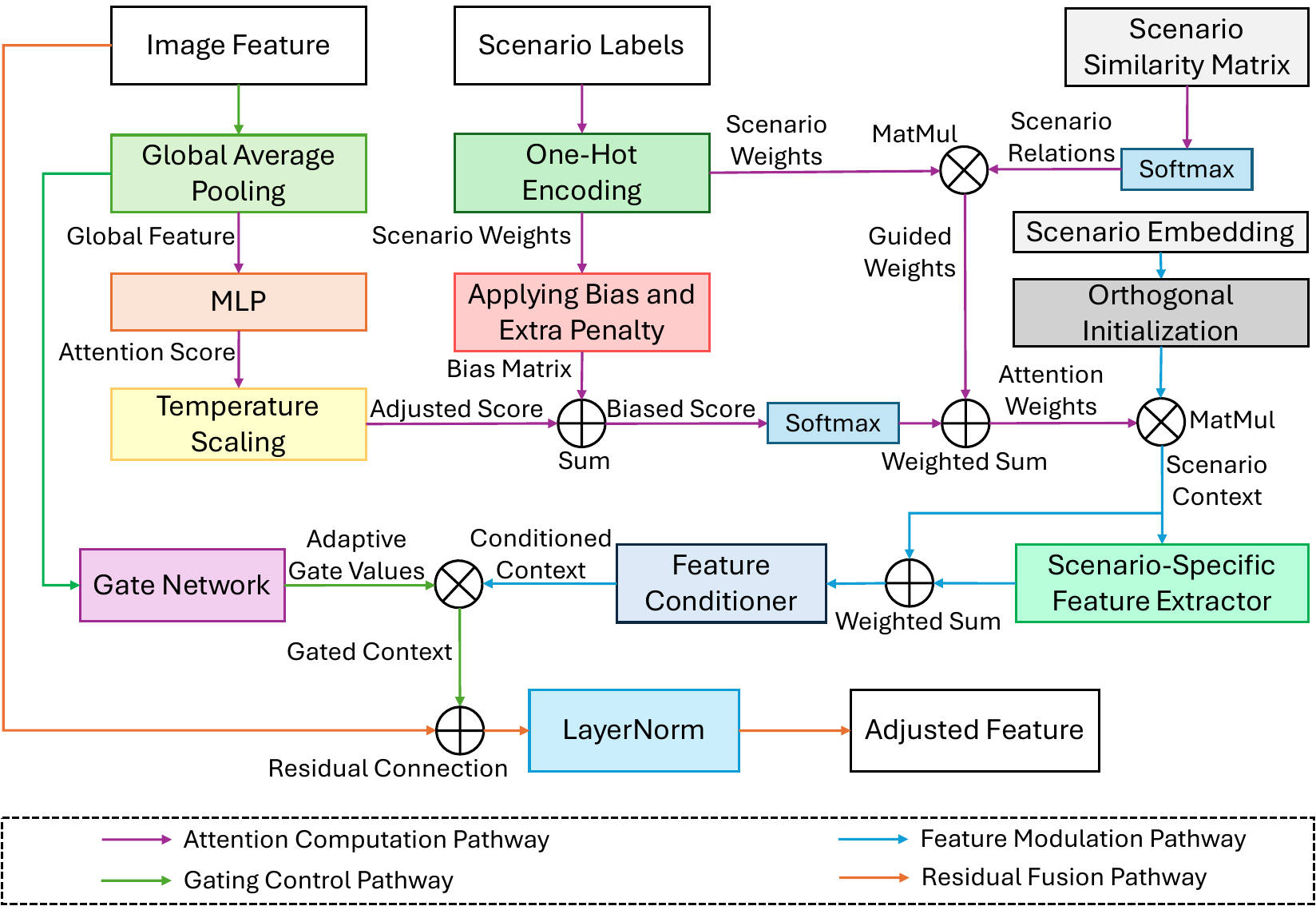}
    \caption{Architecture of the Gated Multi-Scenario Adaptive Attention (GMSAA) module. Attention weights are computed from global image features and blended with similarity-guided supervision. The resulting context is refined by scenario-specific extractors and adaptively gated before fusion with the original token stream.}
    \label{fig:gmsaa}
\end{figure}

To enhance robustness under domain shifts—particularly long-tail scenarios such as snow and fog—we introduce a GMSAA module that conditions visual features on scenario-specific priors. This design injects dynamic attention signals derived from both global scene features and structured scenario relations into the visual backbone, enabling context-aware feature calibration under diverse environments. Figure~\ref{fig:gmsaa} illustrates the detailed architecture.

Let \( z \in \mathbb{R}^{B \times T \times d} \) be the batch of fused visual token embeddings, where \( B \) is the batch size, \( T \) is the token length, and \( d \) is the feature dimension. We begin by summarizing each sample into a global descriptor via temporal average pooling:
\begin{equation}
\bar{z} = \frac{1}{T} \sum_{t=1}^{T} z_{:,t,:} \in \mathbb{R}^{B \times d}.
\end{equation}

These global vectors are passed through a multi-layer perceptron (MLP) to produce unnormalized attention logits over \( C \) scenario types:
\begin{equation}
\mathbf{a} = \phi(\bar{z}) = W_3 \cdot \text{ReLU}\left(W_2 \cdot \text{ReLU}(W_1 \cdot \bar{z})\right) \in \mathbb{R}^{B \times C},
\end{equation}
where \( W_1 \in \mathbb{R}^{d \times d'} \), \( W_2 \in \mathbb{R}^{d' \times d''} \), \( W_3 \in \mathbb{R}^{d'' \times C} \) are trainable projection matrices.

To enhance scenario discrimination, the logits are first temperature-scaled and then enhanced by a scenario-specific self-attention bias vector \( \boldsymbol{\beta}_{\text{self}} \in \mathbb{R}^{C} \). Given scenario label \( d \in \{0,1,2\} \), the adjusted attention becomes:
\begin{equation}
\mathbf{a}' = \frac{\mathbf{a}}{\tau}
            + \boldsymbol{\beta}_{\text{self}} \odot \mathbf{e}_{d}
            + \gamma_{\text{neg}}(d),
\end{equation}
where \(\tau>0\) is a temperature hyper-parameter, 
\(\mathbf{e}_{d}\!\in\!\{0,1\}^{C}\) is the one-hot indicator of the ground-truth scenario $d$, and \(\odot\) denotes Hadamard multiplication broadcast across the \(C\) channels. The vector \(\gamma_{\text{neg}}(d)\!\in\!\mathbb{R}^{C}\) implements a cross-domain penalty:
\[
\gamma_{\text{neg}}(d)=
\begin{cases}
-\,\mathbf{e}_{2}, & d=1\;\text{(Snow)}\\
-\,\mathbf{e}_{1}, & d=2\;\text{(Foggy)}\\
\mathbf{0}, & d=0\;\text{(Normal)},
\end{cases}
\]
so that a value of \(-1\) is added to the snow–foggy logit while normal entries remain unchanged.

To inject scenario supervision, we introduce a learnable scenario–similarity matrix \( \mathbf{S} \in \mathbb{R}^{C \times C} \). A row-wise softmax converts its unnormalized entries into a probabilistic prior, as shown below:
\begin{equation}
\mathbf{S}' = \text{Softmax}(\mathbf{S}), \quad \text{with} \quad
 S'_{ij} = \frac{\exp(S_{ij})}{\sum_{k=1}^{C} \exp(S_{ik})}.
\end{equation}
We then compute a guided attention vector:
\begin{equation}
\mathbf{w}_{\text{guided}} = \mathbf{S}' \cdot \mathbf{e}_{d} \in \mathbb{R}^{C},
\end{equation}
and blend it with the softmax-normalized logits:
\begin{equation}
\mathbf{w} = (1 - \lambda) \cdot \text{Softmax}(\mathbf{a}') + \lambda \cdot \mathbf{w}_{\text{guided}},
\end{equation}
where \( \lambda \) controls the contribution of the learned and prior-guided attention,

This scenario-aware weight vector \( \mathbf{w} \in \mathbb{R}^{B \times C} \) attends to a learned scenario embedding matrix \( \mathbf{E} \in \mathbb{R}^{C \times d} \), initialized with scaled orthogonal weights:
\begin{equation}
\mathbf{c} = \mathbf{w} \cdot \mathbf{E} \in \mathbb{R}^{B \times d}.
\end{equation}
To enhance feature separability, this context is refined by a scenario-specific extractor \( \mathcal{G}_d \), selected based on ground-truth label:
\begin{equation}
\mathbf{c}_d = \mathcal{G}_d(\mathbf{c}) \in \mathbb{R}^{B \times d},
\end{equation}
where \( \mathcal{G}_d \) includes domain-specific MLPs for snow and fog cases and identity mapping for normal conditions. 

Next, a gating mechanism adaptively modulates the impact of the scenario-enhanced features. We compute gate values via a gating network \( \psi : \mathbb{R}^{d} \!\rightarrow\! \mathbb{R}^{1} \):
\begin{equation}
g \;=\; \sigma\!\bigl(\psi(\bar{z})\bigr)\;\in\;\mathbb{R}^{B \times 1},
\end{equation}
where \( \sigma \) denotes the element-wise sigmoid,  and the gating network $\psi$ implemented as a lightweight 
two-layer MLP. 

The gated context is processed by a shared conditioner \( \mathcal{C}: \mathbb{R}^{d} \rightarrow \mathbb{R}^{d} \) and broadcast across tokens before residual fusion:
\begin{equation}
\widetilde{z} = \text{LN}\left(z + g \odot \mathcal{C}(\mathbf{c}_d)\right),
\end{equation}
where \( \odot \) denotes broadcasted element-wise multiplication and LN denotes Layer Normalization.

This mechanism enables GMSAA to adaptively recalibrate visual tokens based on both data-driven and structured scenario priors. It is jointly optimized during training and incurs negligible overhead during inference, supporting robust domain adaptation under adverse or rare driving conditions.

\subsection{Multi-Task Scenario-Aware Contrastive Learning}
\label{sec:multi_task_contrastive}

\begin{figure}[t]
    \centering
    \includegraphics[width=0.95\linewidth]{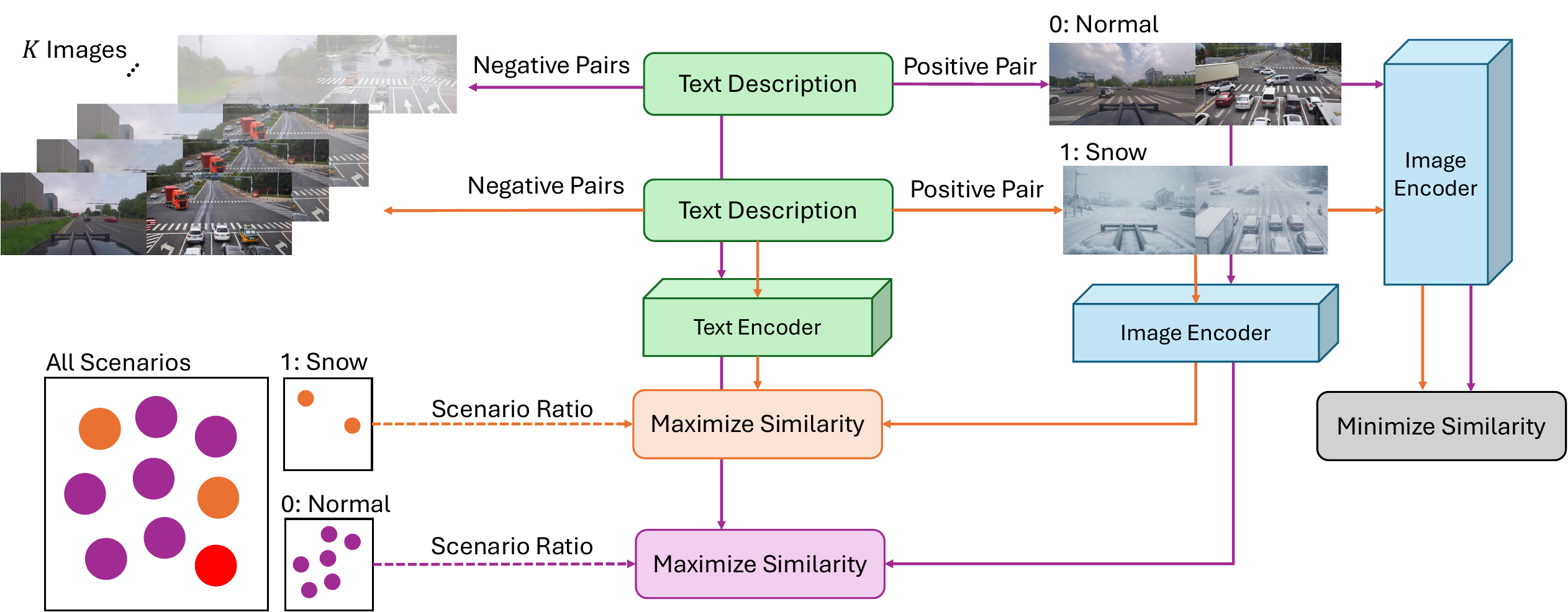}
    \caption{Multi-Task Scenario-Aware Contrastive Learning (MSCL). Image-text pairs are encoded and optimized with cross-modal similarity. Simultaneously, same-scenario image embeddings are encouraged to cluster, guided by scenario frequency-aware weighting.}
    \label{fig:mscl}
\end{figure}

To further improve generalization in long-tail domains, we introduce a MSCL objective that jointly optimizes cross-modal alignment and domain discrimination. As illustrated in Figure~\ref{fig:mscl}, MSCL extends the standard contrastive objective in V2X-VLM (\cite{you2024v2x}) by incorporating scenario-level supervision to explicitly shape the learned embedding space across different scenarios such as normal, snow, and fog scenarios.

Let \( v_i, h_i \in \mathbb{R}^{d} \) denote the \(L_2\)-normalized visual and textual embeddings for the \(i\)-th image-text pair, extracted from the image encoder \( \mathcal{E}_I \) and text encoder \( \mathcal{E}_T \), respectively. We compute the modality contrastive similarity matrix:
\begin{equation}
    S_{ij} = \frac{v_i^\top h_j}{\tau_{\text{mod}}},
\end{equation}
where \( \tau_{\text{mod}} \) is the temperature parameter. The modality contrastive loss follows V2X-VLM (\cite{you2024v2x}) but includes instance-level weighting which will be introduced later:
\begin{equation}
    \mathcal{L}_{\text{mod}} = \sum_{i=1}^B \omega_i \cdot \left[ -\log \frac{\exp(S_{ii})}{\sum_{j=1}^B \exp(S_{ij})} \right],
\end{equation}
where \( B \) is the batch size, and \( \omega_i \) is the weight for the \( i \)-th sample.

To inject scenario awareness, we introduce an intra-modal scenario contrast loss. Let \( d_i \in \{0, 1, 2\} \) be the scenario label of sample \( i \), and define the domain similarity indicator:
\begin{equation}
    \delta_{ij} = \mathbb{1}[d_i = d_j].
\end{equation}
The similarity between image embeddings is computed as:
\begin{equation}
    R_{ij} = \frac{v_i^\top v_j}{\tau_d},
\end{equation}
with a separate temperature \( \tau_d \). The scenario contrastive loss encourages same-scenario samples to cluster while pushing others apart, with pairwise weighting:
\begin{equation}
    \mathcal{L}_{\text{scenario}} = \frac{1}{Z} \sum_{i \neq j} \omega_i \omega_j \cdot \left( \tanh(R_{ij}) - (2\delta_{ij} - 1) \right)^2,
\end{equation}
where \( Z = \sum_{i \neq j} \omega_i \omega_j \) is a normalization constant.

We combine both objectives into a unified multi-task loss:
\begin{equation}
    \mathcal{L}_{\text{mscl}} = \mathcal{L}_{\text{mod}} + \lambda_d \cdot \mathcal{L}_{\text{scenario}},
\end{equation}
where \( \lambda_d \in [0,1] \) balances modality alignment and scenario discrimination.

To simulate and address the inherent scenario imbalance in real-world data, we incorporate instance-level weighting as mentioned above. Let \( f_{d} \) denote the frequency of scenario \( d \) in the current batch. The sample weight for sample \( i \) with label \( d_i \) is:
\begin{equation}
    \omega_i = \frac{1}{f_{d_i}} \left/ \sum_{j=1}^{B} \frac{1}{f_{d_j}} \right..
\end{equation}
These weights are applied to both contrastive objectives to emphasize rare long-tail conditions like snow and fog, preventing the model from underfitting minority domains and ensuring fair representation learning.

The overall formulation enables SEAL to learn robust and semantically aligned features while maintaining domain discriminability and a crucial capability under long-tail and adverse conditions.

%--------------------------------------------------------
\subsection{Training Objective}
\label{sec:training_objective}

The training of SEAL is formulated as a multi-objective optimization problem that jointly supervises cross-modal generation, multi-task scenario-aware contrastive learning, and knowledge distillation from a stronger teacher model. This joint training paradigm ensures both general semantic grounding and robust scenario adaptation.

Let \(\mathcal{L}_{\text{gen}}\) denote the language modeling loss, \(\mathcal{L}_{\text{mscl}}\) the multi-task scenario-aware contrastive loss defined in Section~\ref{sec:multi_task_contrastive}, and \(\mathcal{L}_{\text{kd}}\) the knowledge distillation loss. The overall training objective is:
\begin{equation}
\mathcal{L}_{\text{total}} = \mathcal{L}_{\text{gen}} + \alpha \cdot \mathcal{L}_{\text{mscl}} + \beta \cdot \mathcal{L}_{\text{kd}},
\end{equation}
where \(\alpha\) and \(\beta\) are weighting coefficients that control the contributions of contrastive and distillation objectives.

The generation loss \(\mathcal{L}_{\text{gen}}\) is defined as the cross-entropy between the predicted and ground-truth tokens over autoregressive decoding:
\begin{equation}
\mathcal{L}_{\text{gen}} = - \sum_{t=1}^{T} \log P_{\theta}(y_t \mid y_{<t}, x),
\end{equation}
where \(x\) is the multimodal input, and \(y_t\) is the ground-truth token at step \(t\).

For knowledge distillation, we follow the formulation in V2X-VLM (\cite{you2024v2x}) to align the logits of the student model with those of the teacher model under a softened probability distribution. Let \(\mathbf{z}^{(s)}\) and \(\mathbf{z}^{(t)}\) be the student and teacher logits, respectively, and \(\tau_{\text{kd}}\) the distillation temperature. The distillation loss is given by:
\begin{equation}
\mathcal{L}_{\text{kd}} = \tau_{\text{kd}}^2 \cdot \text{KL}\left( \text{Softmax}\left( \frac{\mathbf{z}^{(t)}}{\tau_{\text{kd}}} \right) \, \| \, \text{Softmax}\left( \frac{\mathbf{z}^{(s)}}{\tau_{\text{kd}}} \right) \right),
\end{equation}
which encourages the student to mimic the teacher's output distribution, improving generalization and stability.

During training, each component contributes distinct regularization and supervision effects:
\begin{itemize}
    \item \(\mathcal{L}_{\text{gen}}\) ensures text output fidelity;
    \item \(\mathcal{L}_{\text{mscl}}\) injects domain-awareness into the embedding space, improving semantic robustness across scenarios;
    \item \(\mathcal{L}_{\text{kd}}\) transfers soft guidance from a more capable teacher model, facilitating better convergence and reducing overfitting on underrepresented cases.
\end{itemize}

This multi-objective training strategy allows SEAL to maintain high performance across both common and rare driving scenarios while preserving semantic alignment and safety-critical awareness.

\section{Experiments}
\label{sec:experiments}
\subsection{Dataset and Evaluation Setup}
\label{sec:dataset}

We evaluate SEAL on the DAIR-V2X dataset (\cite{yu2022dair}), a comprehensive resource designed for V2X cooperative autonomous driving research. The dataset contains 22{,}325 frames from vehicle-side sensors and 10{,}084 frames from infrastructure-side sensors, capturing synchronized RGB images and LiDAR point clouds at up to 25 Hz. Each scene provides time-synchronized multi-view data essential for trajectory prediction, cooperative perception, and sensor fusion tasks.

To address the scarcity of long-tail scenarios in DAIR-V2X—such as extreme weather conditions and visually degraded scenes—we augment the original dataset using the prompt-driven long-tail scene generation pipeline introduced in Section~\ref{sec:realm-longtail}. Specifically, we synthesize photorealistic variations in snow and fog using GPT-4o, conditioned on carefully engineered prompts. This process yields a rich set of dual-view images under rare conditions, each paired with a synthetic language description \( E \) generated by a pre-trained VLM and refined through human curation to ensure semantic accuracy (\cite{you2024v2x}). The resulting triplets \((\widetilde{I}_v, \widetilde{I}_i, E)\) are annotated with scenario labels \(d \in \{0{:}\text{Normal}, 1{:}\text{Snow}, 2{:}\text{Fog}\}\), serving as inputs to the multimodal training and evaluation pipeline described in Section~\ref{sec:realm-overview}.

\begin{figure}[t]
    \centering
    \includegraphics[width=\linewidth]{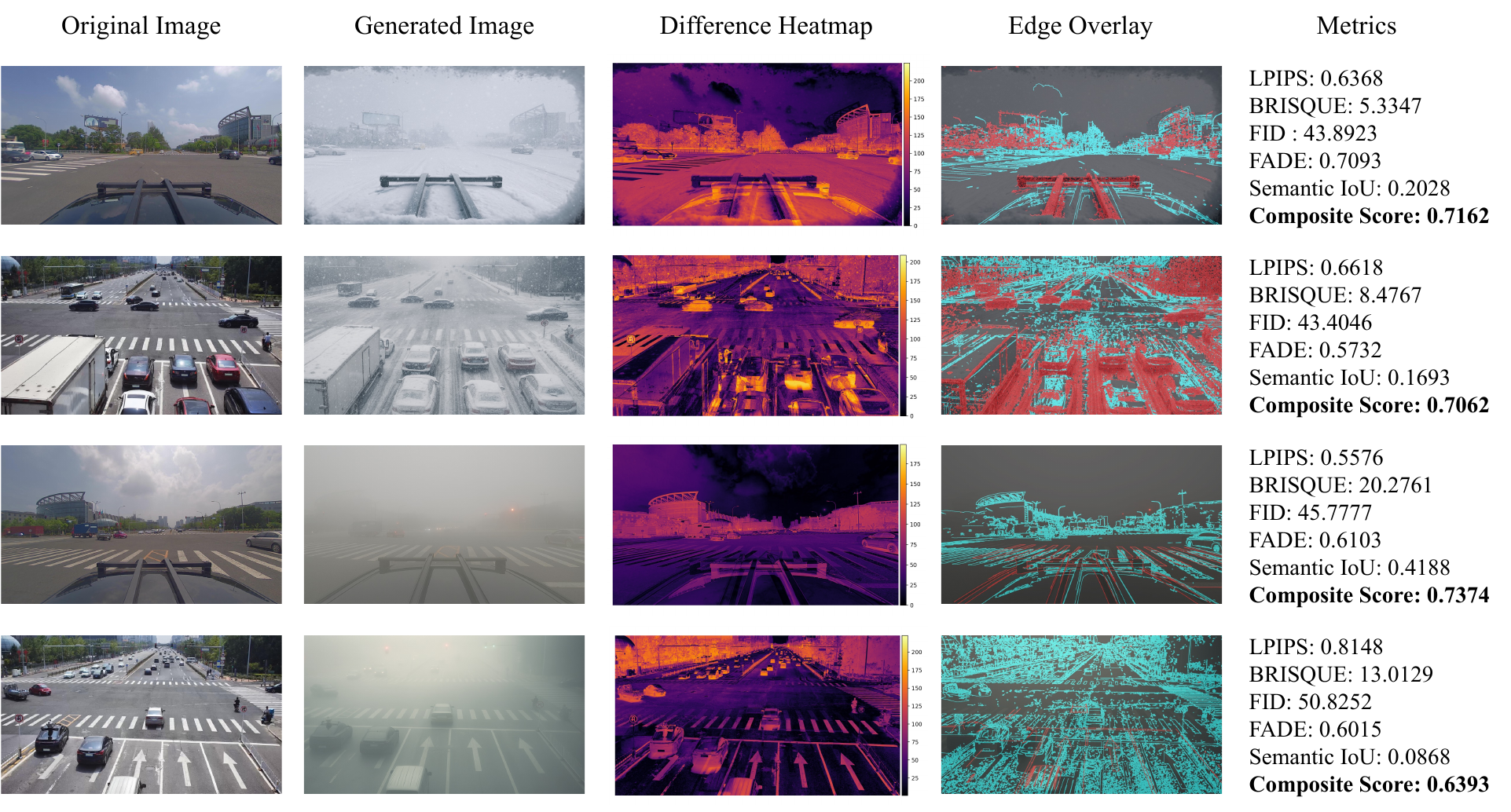}
    \caption{Examples of long-tail data generation and quality assessment. For each pair of original and transformed scenes (snow or fog), we visualize the generated images, scene semantics, edge overlays, and corresponding composite metric scores. These quantitative evaluations validate the fidelity and diversity of the generated data across long-tail conditions.}
    \label{fig:longtail_examples}
\end{figure}

To ensure the quality and realism of these generated scenes, we apply the multi-metric evaluation framework detailed in Section~\ref{sec:realm-longtail}, which includes LPIPS, BRISQUE, FID, FADE, and Semantic IoU. A composite score is derived for each scene using weighted geometric fusion of normalized metric values, designed to holistically reflect perceptual fidelity, visibility degradation, and semantic consistency under each weather condition. Examples of these augmented scenes, along with metric-based assessment results, are visualized in Figure~\ref{fig:longtail_examples}, illustrating both the perceptual quality and structural alignment of the generated data with their real-world counterparts.

To ensure the quality and realism of these generated scenes, we apply the multi-metric evaluation framework detailed in Section~\ref{sec:realm-longtail}, which includes LPIPS, BRISQUE, FID, FADE, and Semantic IoU. A composite score is derived for each scene using weighted geometric fusion of normalized metric values, where LPIPS, BRISQUE, FID, and FADE are first reversed after normalization to ensure higher values consistently indicate better quality. The weights are weather-specific, as shown in Table~\ref{tab:composite-weights}, to emphasize different aspects under snow or fog. This score holistically reflects perceptual fidelity, visibility degradation, and semantic consistency under each condition. Examples of these augmented scenes, along with metric-based assessment results, are visualized in Figure~\ref{fig:longtail_examples}, illustrating both the perceptual quality and structural alignment of the generated data with their real-world counterparts.

\begin{table}[t]
\centering
\caption{Scenario-Specific Weights for Composite Score Calculation}
\label{tab:composite-weights}
\begin{tabular}{l|cc}
\toprule
\textbf{Metric} & \textbf{Snow} & \textbf{Fog} \\
\midrule
LPIPS $\downarrow$ & 0.30 & 0.20 \\
BRISQUE $\downarrow$ & 0.25 & 0.10 \\
FID $\downarrow$ & 0.20 & 0.25 \\
FADE $\downarrow$ & 0.05 & 0.30 \\
Semantic IoU $\uparrow$ & 0.20 & 0.15 \\
\bottomrule
\end{tabular}
\end{table}

By leveraging this augmented dataset, SEAL is trained and evaluated across both normal and long-tail scenarios. This setup enables rigorous testing of its robustness and generalization to underrepresented but safety-critical driving conditions, under consistent multi-modal input structures and scenario-aware supervision.

\subsection{Implementation Details}
\label{sec:implementation}

SEAL is implemented in PyTorch and trained on a single NVIDIA RTX 4090 GPU. The framework builds on Florence-2 (\cite{xiao2024florence}), a state-of-the-art vision-language foundation model known for its efficient architecture and strong performance across multimodal benchmarks. Specifically, we adopt the Florence-2-large checkpoint as the frozen teacher model and Florence-2-base as the student. To improve training efficiency, the vision encoder of the student remains frozen, while the text encoder, fusion transformer, GMSAA, trajectory decoder, and contrastive heads are updated during training.

Training is converged within 13 epochs using the AdamW optimizer with a fixed learning rate of 1e-6, a batch size of 4, and a linear learning rate scheduler. Mixed precision training is enabled via PyTorch Automatic Mixed Precision (AMP) to reduce memory consumption. Each training instance consists of a synchronized dual-view image and a scene-level language prompt, processed by a shared Florence-2 processor. The model is optimized with three objectives: an autoregressive generation loss for trajectory prediction, a MSCL for modality and domain alignment, and a KL-based knowledge distillation loss that transfers supervision from the frozen teacher model to the student. The distillation process uses temperature scaling ($\tau = 2.0$), and loss components are weighted by $\alpha = 0.2$ and $\beta = 0.5$. To mitigate domain imbalance, all objectives incorporate instance-level weighting based on scenario frequency within each batch.

To complement the general training setup, Table~\ref{tab:hyperparameters} summarizes the internal configurations and key hyperparameters specific to the GMSAA and MSCL modules, which includes the initialization and attention settings for scenario-aware adaptation, temperature values for contrastive learning, and blending coefficients that govern feature fusion and guided attention.

% \begin{table*}[t]
% \centering
% \caption{Detailed Hyperparameters in SEAL}
% \label{tab:hyperparameters}
% % \resizebox{\textwidth}{!}{
% \begin{tabular}{l|l|l}
% \toprule
% \textbf{Component} & \textbf{Parameter} & \textbf{Value} \\
% \midrule
% \textbf{GMSAA} 
% & Scaling Factor for Orthogonal Domain Embedding Initialization & 0.1 \\
% & Self-Attention Bias Vector $\boldsymbol{\beta}_{\text{self}}$ for [Normal, Snow, Fog] & $[2.0,\ 2.5,\ 2.5]$ \\
% & Cross-Attention Bias Term for Snow–Fog Mutual Exclusion & $-1.0$ \\
% & Scenario Similarity Matrix $\mathbf{M}$ Initialization &
% $\begin{bmatrix}
% 1.1 & 0.1 & 0.1 \\
% 0.1 & 1.1 & 0.05 \\
% 0.1 & 0.05 & 1.1
% \end{bmatrix}$ \\
% & Feature Fusion Weight for Scenario-Enhanced Context & 0.7 \\
% & Feature Fusion Weight for Raw Context & 0.3 \\
% & Attention Guidance Blending Coefficient $\lambda$ & 0.85 \\
% & Gating Network $\psi(\cdot)$ Architecture & 2-layer MLP \\
% \midrule
% \textbf{MSCL} 
% & Temperature for Modality Contrast $\tau$ & 0.07 \\
% & Temperature for Scenario Contrast $\tau_d$ & 0.1 \\
% & Scenario Contrast Loss Weight $\lambda_d$ & 0.3 \\
% \bottomrule
% \end{tabular}
% % }
% \end{table*}

\begin{table*}[t]
\centering
\caption{Detailed Hyperparameters in SEAL}
\label{tab:hyperparameters}
\begin{tabular}{l|l|l}
\toprule
\textbf{Component} & \textbf{Parameter} & \textbf{Value} \\
\midrule
\textbf{GMSAA} 
& Scaling factor for orthogonal domain embedding initialization & 0.1 \\
& Self-attention bias vector \( \boldsymbol{\beta}_{\text{self}} \) for [Normal, Snow, Fog] & \( [2.0,\ 2.5,\ 2.5] \) \\
% & Cross-attention penalty between Snow and Fog domains & \(-1.0\) \
& Scenario similarity matrix \( \mathbf{S} \) initialization &
\( \begin{bmatrix}
1.1 & 0.1 & 0.1 \\
0.1 & 1.1 & 0.05 \\
0.1 & 0.05 & 1.1
\end{bmatrix} \) \\
& Attention temperature \( \tau \) for scenario distribution & 0.5 \\
& Attention guidance blending coefficient \( \lambda \) & 0.85 \\
& Feature fusion weight for scenario-enhanced context & 0.7 \\
& Feature fusion weight for raw context & 0.3 \\
% & Gating network \( \psi(\cdot) \) architecture & 2-layer MLP 
\midrule
\textbf{MSCL} 
& Temperature for modality contrast \( \tau_{\text{mod}} \) & 0.07 \\
& Temperature for scenario contrast \( \tau_d \) & 0.1 \\
& Scenario contrast loss weight \( \lambda_d \) & 0.3 \\
\bottomrule
\end{tabular}
\end{table*}

\subsection{Experimental Results}
\label{sec:results}

To rigorously evaluate the proposed SEAL, we conduct both overall and scenario-specific experiments on the augmented DAIR-V2X dataset, which contains multi-view driving scenes across three domains: normal, snow, and fog. This dual-level evaluation allows us to measure not only the average performance but also the model's robustness under long-tail and adverse conditions. We compare our method against three state-of-the-art end-to-end cooperative autonomous driving baselines: V2X-VLM (\cite{you2024v2x}), UniV2X (\cite{yu2025end}), and CooperNaut (\cite{cui2022coopernaut}).

We report three quantitative metrics: L2 error in meters, which measures the average Euclidean distance between predicted and ground-truth trajectories at 2.5s, 3.5s, and 4.5s into the future; collision rate, which calculates the percentage of predicted trajectories that intersect with surrounding 3D obstacles; and communication cost, measured in bits per second (BPS), which reflects the overhead of transmitting infrastructure-side images for collaborative autonomous driving. We also report runtime latency and frame-per-second (FPS) for completeness.

Table~\ref{tab:overall} summarizes the overall performance across all test cases. SEAL achieves the lowest average L2 error of 0.6779 and the lowest average collision rate of 0.0662, outperforming all baselines while maintaining comparable communication cost to V2X-VLM. Compared to V2X-VLM, our method improves trajectory accuracy by 9.6\% and reduces collision rate by 17.5\%. UniV2X exhibits extremely low communication overhead and high FPS through its sparse-dense hybrid transmission mechanism, but its accuracy degrades significantly in multimodal settings due to limited representation richness and lack of adaptive alignment. CooperNaut performs moderately well in average L2 error but suffers from high collision rates and elevated bandwidth usage. In contrast, SEAL provides the best trade-off across all three axes: precision, safety, and bandwidth.

\begin{table*}[t]
\centering
\caption{Overall Scenario Performance Comparison}
\label{tab:overall}
\resizebox{\textwidth}{!}{
\begin{tabular}{l|ccc|ccc|c|c|c}
\toprule
\multirow{2}{*}{Method} & \multicolumn{3}{c|}{L2 Error [m] $\downarrow$} & \multicolumn{3}{c|}{Collision Rate [\%] $\downarrow$} & Comm. Cost [BPS] $\downarrow$ & Latency [ms] $\downarrow$ & FPS $\uparrow$ \\
 & 2.5s & 3.5s & 4.5s & 2.5s & 3.5s & 4.5s &  &  &  \\
\midrule
V2X-VLM & 0.7421 & 0.7433 & 0.7656 & 0.0567 & 0.0851 & 0.0993 & $1.24 \times 10^7$ & 254.74 & 15.39 \\
UniV2X (2024 AAAI) & 2.1279 & 2.5702 & 3.0398 & 0.3121 & 0.3191 & 0.3262 & $8.09 \times 10^5$ & 124.71 & 128.30 \\
CooperNaut (CVPR 2022) & 0.8797 & 1.0765 & 1.2990 & 0.2441 & 0.2979 & 0.3404 & $8.19 \times 10^7$ & 132.61 & 7.54 \\
SEAL (Ours) & \textbf{0.6720} & \textbf{0.6731} & \textbf{0.6885} & \textbf{0.0426} & \textbf{0.0709} & \textbf{0.0851} & $1.24 \times 10^7$ & 255.06 & 15.36 \\
\bottomrule
\end{tabular}
}
\end{table*}

Table~\ref{tab:scenario} reports performance under each scenario at the 4.5-second horizon. SEAL consistently achieves strong results across all three scenarios. In the normal scenario, it delivers the lowest collision rate of 0.0847 and a competitive L2 error of 0.6381, outperforming CooperNaut and V2X-VLM in safety. In snow scenarios, which pose significant visual degradation, SEAL maintains a low L2 error of 0.9138, whereas CooperNaut and UniV2X suffer from severe error inflation, exceeding 3.0 and 19.0 meters, respectively. This demonstrates the effectiveness of the GMSAA and MSCL modules in adapting to long-tail conditions. In fog, SEAL again leads in L2 error and sustains a bounded collision rate, outperforming all others in robustness.

\begin{table*}[t]
\centering
\caption{Scenario-Specific Performance Comparison Over 4.5 Seconds}
\label{tab:scenario}
\resizebox{\textwidth}{!}{
\begin{tabular}{l|cc|cc|cc}
\toprule
\multirow{2}{*}{Method} & \multicolumn{2}{c|}{Normal} & \multicolumn{2}{c|}{Snow} & \multicolumn{2}{c}{Fog} \\
 & L2 Error [m] $\downarrow$ & Collision Rate [\%] $\downarrow$ & L2 Error [m] $\downarrow$ & Collision Rate [\%] $\downarrow$ & L2 Error [m] $\downarrow$ & Collision Rate [\%] $\downarrow$ \\
\midrule
V2X-VLM & 0.7296 & 0.1017 & 1.0824 & 0 & 1.0589 & 0.2000 \\
UniV2X (2024 AAAI) & \textbf{0.4258} & 0.3577 & 19.2718 & 0 & 22.5224 & 0.2222 \\
CooperNaut (CVPR 2022) & 0.9500 & 0.3220 & 3.0627 & 0.5385 & 3.1252 & 0.3000 \\
SEAL (Ours) & 0.6381 & \textbf{0.0847} & \textbf{0.9138} & \textbf{0} & \textbf{0.9842} & \textbf{0.2000} \\
\bottomrule
\end{tabular}}
\end{table*}

These results confirm that SEAL offers not only superior prediction accuracy but also consistent safety guarantees across diverse domains. Unlike baselines that collapse under unseen or rare conditions, our model benefits from scenario-aware adaptation and contrastive alignment, preserving performance even in highly challenging environments.

\subsection{Analysis of Scenario-Aware Adaptation}
\label{sec:analysis}

To further understand how SEAL achieves robustness across diverse driving conditions, we conduct a detailed analysis of the GMSAA module and the impact of scenario-specific alignment in the learned feature space.

\begin{figure}[t]
\centering
\includegraphics[width=0.9\linewidth]{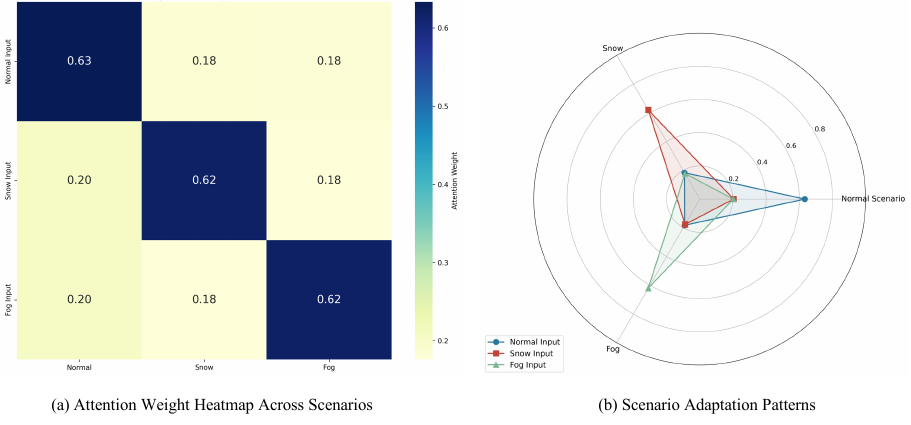}
\caption{Scenario-aware attention learned by GMSAA. Left: attention weight heatmap across domains. Right: adaptation distribution for each input scenario.}
\label{fig:attention}
\end{figure}

Figure~\ref{fig:attention} shows the attention weights across scenarios learned by GMSAA. The left panel presents a heatmap, where each row corresponds to the input scenario and each column to the attended scenario-specific representation. Diagonal entries dominate the matrix (e.g., 0.63 for normal-normal, 0.62 for snow-snow), suggesting the model learns to attend primarily to same-domain knowledge. Moderate off-diagonal values indicate the model also supports cross-domain transfer, which is especially beneficial under adverse conditions with degraded perception quality.

% The right panel of Figure~\ref{fig:attention} presents a radar chart that visualizes the overall adaptation pattern per input scenario. Notably, snow and fog inputs distribute their attention more broadly across domains, while normal inputs are highly focused. This pattern highlights the GMSAA module’s ability to dynamically reallocate attention based on environmental uncertainty, leaning more on cross-domain signals when the input domain is noisy, and specializing when it is reliable.

The right panel of Figure~\ref{fig:attention} visualizes the adaptation pattern for each input scenario using a radar chart. Each input exhibits a highly focused attention distribution, with normal, snow, and fog inputs assigning the majority of their attention to their respective scenario domains. This pattern demonstrates that the GMSAA module effectively learns scenario-aware attention behaviors, assigning dominant weights to the most relevant domain.

% To examine the impact of GMSAA on feature representation, we visualize the feature space before and after adaptation using PCA and t-SNE, shown in Figure~\ref{fig:feature}. Before adaptation (left), features from different scenarios are entangled, especially between fog and normal samples. Post adaptation (right), feature clusters become compact and well-separated.

\begin{figure}[t]
\centering
\includegraphics[width=0.9\linewidth]{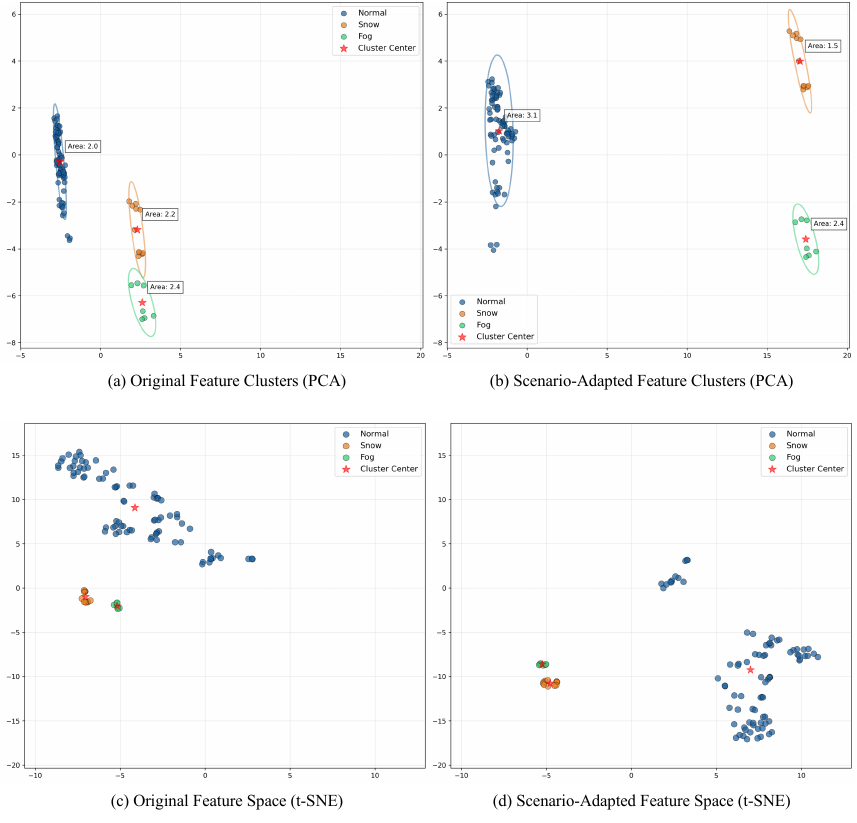}
\caption{Latent feature space before and after GMSAA adaptation. Left: original features. Right: adapted features with clearer domain separation.}
\label{fig:feature}
\end{figure}

To examine the impact of GMSAA on feature representation, we visualize the feature space before and after adaptation using PCA and t-SNE, as shown in Figure~\ref{fig:feature}. Before adaptation (left), scenario-wise clusters exhibit partial overlap and loose compactness, especially between fog and snow samples. After applying GMSAA (right), the representations become significantly more compact and better separated, with clearer inter-scenario margins and more concentrated cluster centers.

Specifically, PCA projections in the top row show that scenario-adapted features exhibit reduced intra-class variance and better inter-class margins. Similarly, t-SNE embeddings in the bottom row reveal that fog and snow clusters become more isolated from normal clusters after GMSAA. This improved separation demonstrates GMSAA’s capacity to disentangle overlapping features, thus promoting semantic clarity and robustness across long-tail scenarios.

In summary, the analysis validates the GMSAA module’s dual contribution: (1) it allocates attention asymmetrically to compensate for uncertain modalities, and (2) it improves the semantic structure of learned features to support robust performance under diverse and long-tail driving scenarios.

\subsection{Ablation Study}
\label{sec:ablation}

To understand the contribution of each component in SEAL, we perform an ablation study along three dimensions: (1) general input modalities and major modules, (2) internal designs of the GMSAA module, and (3) the architectural factors in the MSCL objective. The evaluation is conducted using average L2 error over the prediction horizon of 2.5, 3.5, and 4.5 seconds across all scenarios. The results are summarized in Tables~\ref{tab:ablation-general}--\ref{tab:ablation-mscl}.

\subsubsection{General Modality and Module Contributions}
\label{sec:ablation-general}

We begin by disabling each key input or architectural module in isolation to examine its influence on overall performance. Each ablation case is constructed as follows:

\begin{itemize}
    \item \textbf{w/o Infrastructure Images}: Removes infrastructure-side views from the input, leaving the model with vehicle-side images solely.
    \item \textbf{w/o Scene Description}: Removes the scene description, disabling text-based contextual reasoning.
    \item \textbf{w/o GMSAA}: Bypasses the GMSAA module by feeding original image features directly into the transformer encoder without injecting scenario-aware priors from scenario labels.
    \item \textbf{w/o MSCL}: Disables the entire scenario-aware and multimodel contrastive loss and relies only on generation and distillation supervision.
\end{itemize}

As shown in Table~\ref{tab:ablation-general}, the most dramatic degradation comes from removing infrastructure images, increasing the average L2 error from 0.6779 to 2.9347. This emphasizes the necessity of collaborative perception in cooperative driving, compared with single-vehicle intelligence. Scene description also provides meaningful gains, while both GMSAA and MSCL individually offer sizable accuracy boosts, which validates the need for adaptive attention and representation alignment.

\begin{table}[t]
\centering
\caption{General Module Ablation Results}
\label{tab:ablation-general}
% \resizebox{\linewidth}{!}{
\begin{tabular}{l|ccc|c}
\toprule
\multirow{2}{*}{Method} & \multicolumn{3}{c|}{L2 Error [m] $\downarrow$} & Avg. \\
 & 2.5s & 3.5s & 4.5s & \\
\midrule
w/o Infrastructure Images & 2.9259 & 2.9277 & 2.9507 & 2.9347 \\
w/o Scene Description & 2.1951 & 2.1963 & 2.2254 & 2.2057 \\
w/o GMSAA & 1.4774 & 1.4785 & 2.2061 & 1.7745 \\
w/o MSCL & 0.8120 & 0.8189 & 1.5421 & 1.0577 \\
\textbf{SEAL} & \textbf{0.6720} & \textbf{0.6731} & \textbf{0.6885} & \textbf{0.6779} \\
\bottomrule
\end{tabular}
% }
\end{table}

\subsubsection{Design Analysis of Gated Multi-Scenario Adaptive Attention}
\label{sec:ablation-gmsaa}

Subsequently, we dissect the GMSAA module by ablating individual subcomponents to assess their contributions. Each variant is constructed as follows:

\begin{itemize}
    \item \textbf{w/o Orthogonal Initialization}: Replaces the orthogonal initialization of domain embeddings with standard Gaussian initialization, reducing the separability among domain-specific representations.
    \item \textbf{w/o Snow-Fog Scenario Separation}: Removes the explicit separation between snow and fog scenarios by disabling the mutual exclusion bias in the attention logits, weakening the model’s ability to distinguish between these two long-tail conditions.
    \item \textbf{w/o Scenario Similarity Guidance}: Removes the domain similarity matrix and guided attention mechanism, relying solely on learned attention weights without incorporating inter-scenario relational priors.
    \item \textbf{w/o Scenario-Specific Feature Extractors}: Replaces all scenario-specific extractors with a single shared identity mapping, eliminating the specialized processing paths for snow and fog and thereby removing their differentiated treatment during feature conditioning.
    \item \textbf{w/o Adaptive Gating Control}: Removes the gating mechanism and directly uses the conditioned features without modulation by the global gate values, thereby disabling dynamic adaptation based on scene-level context.
\end{itemize}

As shown in Table~\ref{tab:ablation-gmsaa}, the removal of scenario-specific feature extractors results in the most dramatic degradation, increasing the average L2 error to 2.2182. This highlights the critical role of dedicated processing paths for snow and fog in capturing their unique visual and structural characteristics, which are easily lost when using a generic representation. Disabling the adaptive gating control also causes a significant performance drop, indicating that dynamically modulating feature importance based on global context is essential for balancing information flow across domains. Furthermore, removing scenario similarity guidance slightly inflates the error, suggesting that even subtle priors from domain relationships enhance attention alignment, especially under limited training signals for long-tail scenarios. Overall, each subcomponent of the GMSAA module provides complementary benefits, with the full design yielding the most robust and domain-adaptive trajectory planning.

\begin{table}[t]
\centering
\caption{Ablation Study for Gated Multi-Scenario Adaptive Attention}
\label{tab:ablation-gmsaa}
% \resizebox{\linewidth}{!}{
\begin{tabular}{l|ccc|c}
\toprule
\multirow{2}{*}{Method} & \multicolumn{3}{c|}{L2 Error [m] $\downarrow$} & Avg. \\
 & 2.5s & 3.5s & 4.5s & \\
\midrule
w/o Orthogonal Initialization & 0.7694 & 0.7710 & 0.8139 & 0.7848 \\
w/o Snow-Fog Scenario Separation & 0.7671 & 0.7687 & 0.7995 & 0.7784 \\
w/o Scenario Similarity Guidance & 0.7766 & 0.7782 & 0.8130 & 0.7893 \\
w/o Scenario-Specific Feature Extractors & 2.1954 & 2.2262 & 2.2330 & 2.2182 \\
w/o Adaptive Gating Control & 0.7557 & 0.7573 & 1.4922 & 1.0017 \\
\textbf{SEAL} & \textbf{0.6720} & \textbf{0.6731} & \textbf{0.6885} & \textbf{0.6779} \\
\bottomrule
\end{tabular}
% }
\end{table}

\subsubsection{Design Analysis of Multi-Task Scenario-Aware Contrastive Learning}
\label{sec:ablation-mscl}

Lastly, we ablate the three core designs within the MSCL module to validate their necessity. The configurations are described below:

\begin{itemize}
    \item \textbf{w/o Scenario Weighting}: Treats all training instances equally regardless of domain prevalence or sample rarity.
     \item \textbf{w/o Scenario Awareness}: Removes the scenario-level contrastive loss, eliminating scenario-aware supervision that promotes separability between different scenarios in the embedding space.
    \item \textbf{w/o Text-Image Discrimination}: Removes the cross-modal contrastive loss that aligns visual and textual embeddings, which weakens the model's ability to effectively fuse multimodal information.
\end{itemize}

As shown in Table~\ref{tab:ablation-mscl}, removing the text-image discrimination objective results in the largest performance degradation, with the L2 error increasing to 1.3895. This highlights the central role of cross-modal alignment in SEAL, as it enables the model to effectively integrate semantic priors from scene descriptions with visual cues. Removing the scenario weighting mechanism also leads to a substantial performance drop, demonstrating the necessity of calibrating gradient contributions based on domain frequency to mitigate long-tail bias. Finally, ablating the scenario-aware contrastive loss moderately impairs performance, underscoring its role in enforcing inter-domain separability and enhancing robustness under domain shift. These findings validate that each MSCL component contributes uniquely—through balancing, separation, and alignment—to the model’s multimodal reasoning capability.

\begin{table}[t]
\centering
\caption{Ablation Study for Multi-Task Scenario-Aware Contrastive Learning}
\label{tab:ablation-mscl}
% \resizebox{\linewidth}{!}{
\begin{tabular}{l|ccc|c}
\toprule
\multirow{2}{*}{Method} & \multicolumn{3}{c|}{L2 Error [m] $\downarrow$} & Avg. \\
 & 2.5s & 3.5s & 4.5s & \\
\midrule
w/o Scenario Weighting & 0.7613 & 0.7624 & 1.4988 & 1.0075 \\
w/o Scenario Awareness & 0.7220 & 0.7236 & 0.7577 & 0.7344 \\
w/o Text-Image Discrimination & 0.8106 & 0.8122 & 2.5457 & 1.3895 \\
\textbf{SEAL} & \textbf{0.6720} & \textbf{0.6731} & \textbf{0.6885} & \textbf{0.6779} \\
\bottomrule
\end{tabular}
% }
\end{table}

\subsection{Runtime and Qualitative Evaluation}
\label{sec:runtime-qualitative}

To further assess the practical applicability of SEAL, we analyze both its computational runtime and qualitative planning performance across diverse scenarios. These evaluations offer complementary insights into the model’s efficiency and real-world robustness.

\paragraph{Runtime Analysis.}
Table~\ref{tab:runtime} presents the breakdown of runtime latency across key stages of the inference pipeline. Preprocessing—which includes tokenization, resizing, and image normalization—accounts for the majority of total latency at 70.01\%, followed by model inference at 28.99\%. The actual forward pass through the VLM-based architecture is relatively efficient, with a latency of 73.93 ms. Postprocessing and residual overhead, which include decoding and synchronization steps, contribute marginally together under 3\%. Overall, SEAL operates at a stable frame rate of 15.36 FPS, meeting real-time processing requirements for autonomous driving applications.

\begin{table}[t]
\centering
\caption{Runtime Evaluation}
\label{tab:runtime}
\resizebox{\linewidth}{!}{
\begin{tabular}{l|l|c|c}
\toprule
\textbf{Process} & \textbf{Description} & \textbf{Latency (ms)} & \textbf{Proportion (\%)} \\
\midrule
Preprocessing & Tokenization and image processing & 178.56 & 70.01 \\
Inference & Forward pass through the model & 73.93 & 28.99 \\
Postprocessing & Decoding the model outputs & 1.38 & 0.54 \\
Residual Overhead & Minor operations, such as data loading, synchronization, and loop overhead & 4.56 & 1.79 \\
\midrule
Total &  & 255.06 & 100.00 \\
FPS &  & 15.36 & - \\
\bottomrule
\end{tabular}}
\end{table}

\begin{figure*}[t]
\centering
\includegraphics[width=\textwidth]{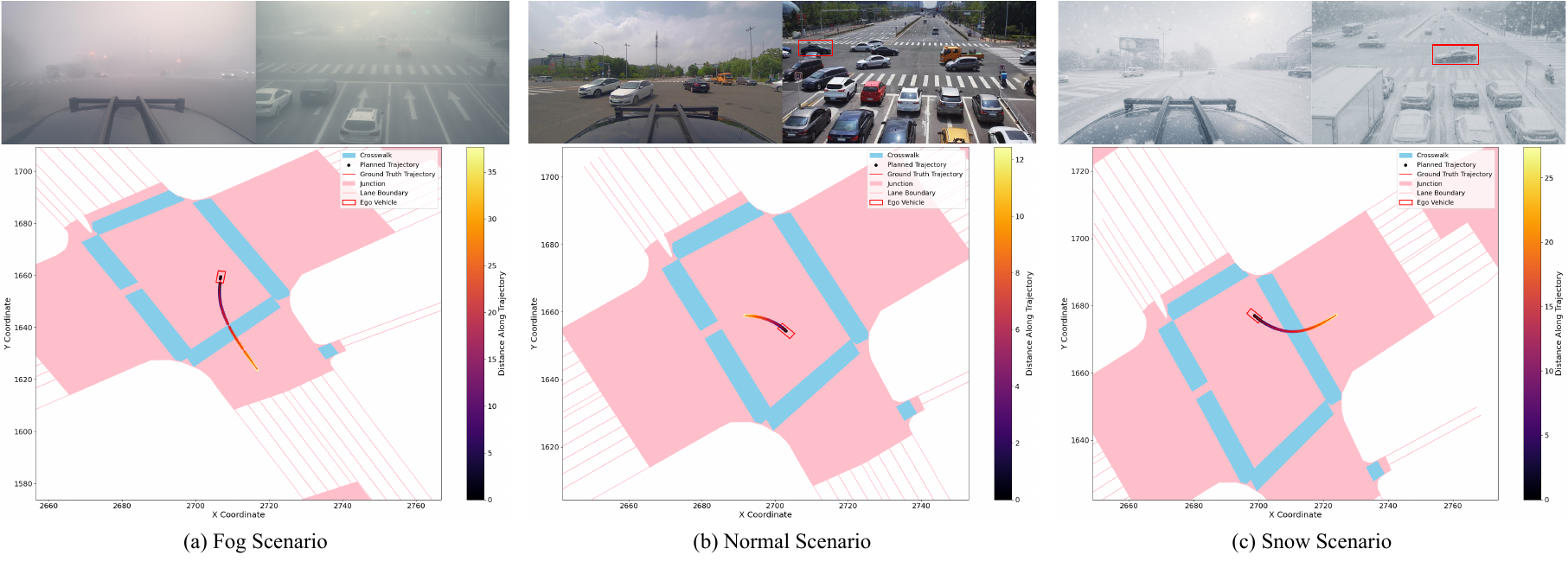}
\caption{
% Qualitative visualization of planned trajectories across fog (a), normal (b), and snow (c) scenarios. Top: multi-view inputs. Bottom: HD maps with planned and ground truth vehicle location.
Qualitative visualization of planned trajectories under fog (a), normal (b), and snow (c) scenarios. The top row shows vehicle-side and infrastructure-side views, where bounding boxes in infrastructure-side images highlight the ego vehicle's position when visible. Bottom row presents the corresponding HD maps with planned trajectories, ground truth paths, and lane semantics. The visualization demonstrates how SEAL maintains accurate and context-aware planning performance across diverse and challenging conditions.}
\label{fig:qualitative}
\end{figure*}

\paragraph{Qualitative Results.}
Figure~\ref{fig:qualitative} visualizes example trajectories generated by SEAL under three different weather conditions: fog, normal, and snow. In each case, the planned trajectory is shown alongside the ground-truth path on a high-definition map. Notably, SEAL accurately predicts smooth and safe trajectories even under heavy snow and dense fog, where visibility is drastically reduced and sensor noise is prominent.

In the fog scenario (Figure~\ref{fig:qualitative}a), the model produces a trajectory that closely follows the true path despite degraded visibility in vehicle- and infrastructure-side views. This highlights the robustness of our GMSAA and MSCL modules in extreme long-tail settings. In contrast, the normal scenario (Figure~\ref{fig:qualitative}b) exhibits minimal deviation and optimal lane alignment, as expected under clear perception conditions. In the snow scenario (Figure~\ref{fig:qualitative}c), SEAL again successfully adjusts to reduced road friction and occlusion, demonstrating its strong scenario adaptation capabilities.

These evaluation results validate the architectural advantages of scenario-aware fusion and contrastive alignment and illustrate SEAL’s ability to generalize across domains while maintaining real-time inference speed.

\section{Conclusion}
\label{sec:conclusion}
Rare, diverse weather conditions pose significant challenges for autonomous driving safety. In this work, we introduced SEAL, a robust end-to-end cooperative autonomous driving framework empowered by VLM and adaptive long-tail modeling. Building upon our prior work, V2X-VLM, SEAL advances the state of the art through three key contributions: (i) a prompt-driven pipeline for generating and evaluating realistic long-tail scenarios via pre-trained foundation models, (ii) a GMSAA module that dynamically injects scenario priors to recalibrate perception under domain shifts, and (iii) a MSCL objective that aligns multimodal embeddings while promoting domain-level feature separability. Together, these components enable SEAL to maintain semantic robustness and planning accuracy across diverse and challenging driving conditions.

Extensive experiments demonstrate that SEAL outperforms existing cooperative end-to-end autonomous driving baselines under both standard and adverse scenarios, especially in snow and fog domains where most prior methods struggle. Detailed ablation and visualization analyses further confirm the effectiveness of each proposed module in improving generalization and long-tail robustness.

Future work includes expanding the coverage of long-tail scenarios to encompass more diverse and safety-critical edge cases, such as unexpected emergencies, rare agent behaviors, and dynamically changing infrastructure conditions. Modeling these challenging events is vital for enhancing generalization and reliability in open-world deployments. Another key direction involves reducing communication overhead in V2X cooperation through techniques such as selective feature sharing, adaptive compression, and efficient token-based fusion, aiming to support scalable and bandwidth-efficient cooperative autonomy. Additionally, incorporating chain-of-thought (CoT) reasoning into the vision-language backbone may further strengthen sequential decision-making and improve interpretability under connected and complex driving contexts.

% \appendix
% \section{My Appendix}
% Appendix sections are coded under \verb+\appendix+.

% \verb+\printcredits+ command is used after appendix sections to list 
% author credit taxonomy contribution roles tagged using \verb+\credit+ 
% in frontmatter.

\printcredits

\bibliographystyle{cas-model2-names}

\bibliography{cas-refs}


% Author biography without author photo.
% Author biography. Author biography. Author biography.
% Author biography. Author biography. Author biography.
% Author biography. Author biography. Author biography.
% Author biography. Author biography. Author biography.
% Author biography. Author biography. Author biography.
% Author biography. Author biography. Author biography.
% Author biography. Author biography. Author biography.
% Author biography. Author biography. Author biography.
% Author biography. Author biography. Author biography.

% \bio{figs/pic1}
% Author biography with author photo.
% Author biography. Author biography. Author biography.
% Author biography. Author biography. Author biography.
% Author biography. Author biography. Author biography.
% Author biography. Author biography. Author biography.
% Author biography. Author biography. Author biography.
% Author biography. Author biography. Author biography.
% Author biography. Author biography. Author biography.
% Author biography. Author biography. Author biography.
% Author biography. Author biography. Author biography.

% \bio{figs/pic1}
% Author biography with author photo.
% Author biography. Author biography. Author biography.
% Author biography. Author biography. Author biography.
% Author biography. Author biography. Author biography.
% Author biography. Author biography. Author biography.

\end{document}